\documentclass{bmvc2k}
\usepackage{mwe}
\usepackage{array,multirow,graphicx}
\usepackage{float}
\usepackage{graphicx}
\usepackage{amsmath,amssymb} 
\usepackage{color}
\usepackage{makecell}
\usepackage{xcolor,colortbl}
\usepackage{cleveref}
\definecolor{Gray}{gray}{0.85}
\definecolor{LightCyan}{rgb}{0.88,1,1}
\newcolumntype{a}{>{\columncolor{Gray}}c}
\newcolumntype{b}{>{\columncolor{LightCyan}}c}
\usepackage{pdfpages}

\title{Mining for meaning: from vision to language through multiple networks consensus}

\addauthor{Iulia Du\c t\u a\textsuperscript{*}}{iduta@bitdefender.com}{12}
\addauthor{Andrei Liviu Nicolicioiu\textsuperscript{*}}{anicolicioiu@bitdefender.com}{13}
\addauthor{Simion-Vlad Bogolin\textsuperscript{*}}{vladbogolin@gmail.com}{4}
\addauthor{Marius Leordeanu}{marius.leordeanu@imar.ro}{34}

\addinstitution{
 Bitdefender, Romania
}
\addinstitution{
 University of Bucharest, Romania
}
\addinstitution{
 University Politehnica of Bucharest, \\ Romania
}
\addinstitution{Institute of Mathematics of the \\ Romanian Academy}

\runninghead{I. Du\c t\u a \protect\etal}{Mining for meaning}

\def\etal{\emph{et al}\bmvaOneDot}

\begin{document}

\maketitle

\begin{abstract}
Describing visual data into natural language is a very challenging task, at the intersection of computer vision, natural language processing and machine learning. Language goes well beyond the description of physical objects and their interactions and can convey the same abstract idea in many ways. It is both about content at the highest semantic level as well as about fluent form. Here we propose an approach to describe videos in natural language by reaching a consensus among multiple encoder-decoder networks. Finding such a consensual linguistic description,
which shares common properties with a larger group, has a better chance to convey the correct meaning. We propose and train several network architectures and use different types of image, audio and video features. Each model produces its own description of the input video and the best one is chosen through an efficient, two-phase consensus process. We demonstrate the strength of our approach by obtaining state of the art results on the challenging MSR-VTT dataset.
\end{abstract}

\section{Introduction}

The task of describing videos into natural language is one of the most exciting and still unsolved problems in artificial intelligence today. Solving this task would help decode many important questions about how the mind works, how we perceive the world, how we think and then communicate to one another. Efficient methods for vision to language translation would also have an immense practical value, with applications in many areas ranging from technology to medicine and entertainment.

The problem is hard to formulate in the traditional supervised machine learning paradigm. For every video sequence, there is, in principle, an infinite number of correct descriptions in natural language. Many leading cognitive scientists, such as Noam Chomsky \cite{chomsky2002syntactic} and Steven Pinker~\cite{pinker2003language} among others, observed that every human utterance is unique. Thus, it is not reasonable to enforce an exact rigid form 
on a video description in natural language.
Vision and language are deeply linked and evolve naturally during early age~\cite{pinker2003language}. Given sufficient training data, one could expect 
an end-to-end deep learning approach to be able to translate vision into language. However, there is a lot of work to do in extracting meaning from language and being able to evaluate linguistic descriptions based on both meaning and form.

In this paper, we present an approach to address these challenges based on finding the consensual linguistic description among multiple vision to language translation models. While each model individually is able to generate well formed sentences that generally obey grammatical rules, it is the consensus among many models that best captures the hidden meaningful content and significantly outperforms the individual models on the tested evaluation metrics. 

\paragraph{Related work.}
Recurrent Neural Networks (RNNs) and Long Short-Term Memory (LSTM) \cite{hochreiter1997long}, which are successful on text generation, are the basis for current models in vision to language translation. Initial works~\cite{venugopalan2014translating} on video captioning using RNNs perform average feature pooling over the video and bring the task closer to image captioning. The strategy works well for short videos, in which a single major event takes place~\cite{yu2016video}. For longer videos, different video encoding schemes are proposed. These schemes use either a recurrent encoder~\cite{venugopalan2015sequence,donahue2015long} or an attention model~\cite{yao2015describing}. In~\cite{yu2016video} authors use a hierarchical RNN model, with a sentence generator and a separate paragraph generator. The sentence decoder has an attention mechanism to focus on video features while exploiting spatial attention.

Methods for selecting captions from multiple models have been proposed in \cite{ruc_uva_dong2016early_ruc_rerank, shetty2016frame}. Unlike our work, they learn a compatibility score between a single sentence and a given video, without taking in consideration the whole group of output sentences. The authors of \cite{chen2017video_topic_guidance} use latent topics to guide the sentence generation process. They mine a number of K topics and implicitly learn an ensemble of K decoders, one for each topic. The number of parameters is reduced by a 3-way factorization~\cite{krizhevsky2010factored_3way} of the mixture of all topic parameters.
 
External data can be used to enlarge the linguistic knowledge~\cite{external_language_data_venugopalan2016improving}. In~\cite{pasunaruRamBnMoh} the authors use additional tasks for improving the learning process: an unsupervised video prediction and a language entailment generation task. The usual way of predicting the next word given the previous correct one using standard LSTM decoders creates a difference between the distributions at training vs. testing time, an issue called exposure bias. To tackle it, reinforcement learning approaches have been studied in the context of image captioning ~\cite{ranzato2015sequence_mixer, Liu_2017_ICCV_policy_spider, chen2017show_adapt}. An already trained model is improved by a policy gradient method that works on whole output sentences, guided by a non-differentiable reward, given by the language metrics. Recently this approach has been applied also for video captioning~\cite{PasRamBanMoh,Wang2017VideoCV}.

\paragraph{Main contributions.}
The main contributions of our approach are:
1) We describe videos in sentences by finding a consensus among multiple encoder-decoder networks. While the individual encoder-decoder networks are able to produce well-formed, fluent sentences it is the consensus among many models that improves the content. The consensus process has two stages. Firstly, we choose a select group of sentences that score well when are evaluated against the others. The next stage we use an Oracle network to pick the final best sentence. The proposed approach achieves state of the art results on the MSR-VTT benchmark.

2) We propose two novel architectures and perform extensive tests with many others adapted from the literature. We also study how different kinds of image, audio or video features influence the final result. We conclude that features that are pretrained on different but related tasks, such as word label prediction or action classification, could impact performance more than the individual architectures. Thus simpler yet higher level tasks such as action or word prediction, could be an effective intermediary between vision and language. 

\section{Network architectures}

We perform tests with different network architectures and types of features as explained in this section. We also propose two novel encoder-decoder networks for video to language translation. All tested models are based on the encoder-decoder paradigm, and they all have the same LSTM decoder structure. They differ only in the way video content is encoded and in the types of features used.

\paragraph{Seq2Seq model:}
Describing a video in language could be formulated as a machine translation~\cite{bahdanau2014neural} problem, but it is much more difficult in practice. Instead of translating sentences into a foreign language, now we have to translate visual features into language. Since videos consist of sequences of frames, it is natural to use a recurrent net such as LSTM to produce the encoding. Thus, for every frame the encoder LSTM receives visual features extracted from that particular frame, together with the previous hidden state. The LSTM output at the last step represents the encoding. The encoding could be augmented with extra contextual information by concatenating different visual or audio features, which could be pretrained for different, but related tasks. Such features encode additional knowledge that brings significant improvement in performances (Section \ref{ablation}). We use LSTM cells with one layer in our experiments. The encoder has 512 hidden dimension and the decoder 
has hidden dimension 512 plus the size of the additional features. Moreover, the encoder part could include an attention mechanism similar to~\cite{bahdanau2014neural} to weight differently the encoder hidden state from each time step, before linearly combining them for the final encoding. In experiments, the attention mechanism brought a marginal improvement.

The decoder, common to all models, works as follows: it starts to output one word at a time. While training, the model receives the previous hidden state and the ground truth word from the previous time step to generate the next word in the sentence. At test time, the ground truth word is replaced by the generated output from the previous state. Our models are trained using softmax cross-entropy loss, unless otherwise specified.

\begin{figure}[t!]
\includegraphics[width=.99\textwidth]{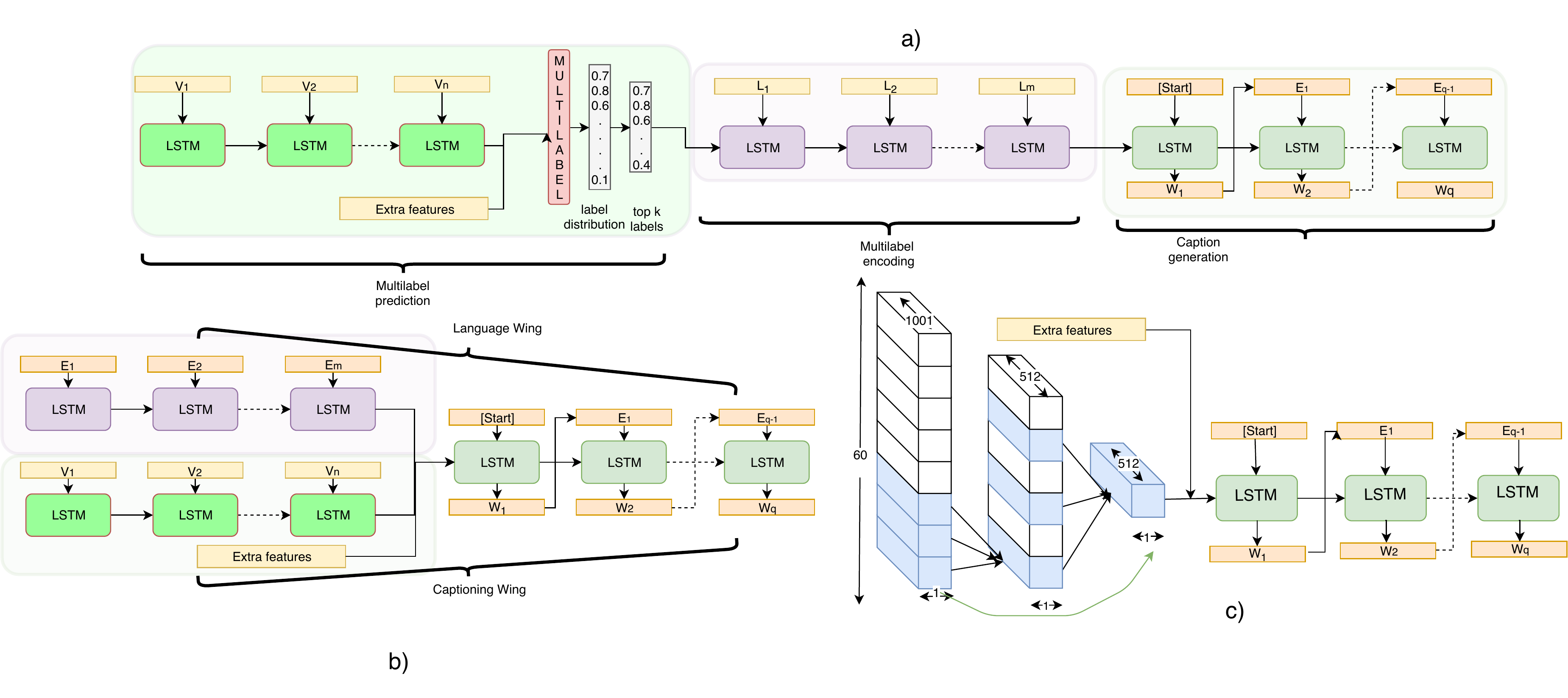}
\caption{Our main architectures: a) Two-Stage: vision to words to sentences, b) Two-Wings network, c) Temporal Convolutional Network (TCN). The architectures differ in structure significantly and generally output different sentences, but have a similar overall performance.}
\label{fig:arhitecturi}
\end{figure}

\paragraph{Two-Wings network with sentence reconstruction:}
The seq2seq model tends to produce, in experiments, simple sentences with very limited vocabulary. We want a stronger decoder, able to capture more realistic, complex sentences. We aim to accomplish this by a model
which we term the \textbf{Two-Wings network} due to its dual language and vision encoder. Besides the video to language pathway the Two-Wings net has a second encoder-decoder branch for language reconstruction (Figure \ref{fig:arhitecturi} b). The decoders of the two networks are shared and the second branch is only used at training time for a stronger decoder with more generalization power. 
The two branches (wings) are trained alternatively, with the decoder having shared parameters. The second, language reconstruction wing is trained for a few iterations and learns to reconstruct broken sentences or to create fluent sentences from sets of words. 
For a given sentence, we randomly remove half of its words and then shuffle them. We have chosen this approach to make the model more robust since without introducing this noise, the model would simply copy the input. In this way the model can benefit from learning how a correct sentence looks like from a huge amount of text data.

The first wing for vision to language translation is trained for a few iterations while using the same decoder as the other one. By forcing a common decoding part we hope to learn a common embedding between language and vision with stronger generalization power, better able to capture meaningful content across vision and language. One advantage of the language reconstruction path is that it can be trained on any freely available dataset of texts, unlike the second path, for which there is limited training data available. 

We train the language reconstruction wing using a set of 10M sentences (of maximum 20 words) extracted from Wikipedia along with training video captions. Note that the language reconstruction model is used only during training in order to learn a more powerful decoder. During testing, only the second video to sentence network is used.

\paragraph{Two-Stage Network, from video to words to sentences:}

The Two-Wings network uses the language reconstruction encoder only to train a stronger sentence decoder. The second model we propose, the Two-Stage net, 
puts two encoder-decoder nets one after the other (Figure \ref{fig:arhitecturi}).
The first stage net learns to output words from videos. The second stage net learns to produce sentences from the sets of words given by the first.
Thus, word labels provide an intermediate semantic interpretation standing between video data and the final sentence. This idea could increase the generalization power, by focusing on content first (as it is captured by individual words), before learning to produce fluent sentences. Note that the words generation is treated as a multi-label classification problem, with no order imposed. 

For the first stage net, we keep the same video encoding scheme as in the Two-Wings model. We replace the language decoder by a model for multi-label prediction, consisting of three fully connected layers that predict the probability of each word label. Given the predicted word labels, we then output a sentence at the second stage, using the same encoder-decoder net used in the language branch of the Two-Wings net.

To form our words labels vocabulary we select the most frequent nouns and verbs from the captioning dataset, resulting in 3059 labels.  During training this part, for every sentence we extract the labels and with a given probability we randomly remove some labels and add other ones for robustness. As in the Two-Wings case, we augment the training set by extracting sentences from Wikipedia that contain some words from the 3059 set of labels.

Initially, the two encoder-decoder stages are learned separately and then finetuned end-to-end on the captioning dataset. To make end-to-end training possible, we keep the whole path, from video input to final sentence differentiable, as follows. For each $j$ in the top K predicted labels, we multiply the label embedding $E(j)$ with its predicted soft probability and obtain a differentiable latent representation $L_t = p_{t,j} * E(j) $. Thus, the gradients could be propagated through $p_{t,j}$ back to the video encoder.

\paragraph{Temporal convolutional network}

In experiments, many of the generated sentences, although fluent, do not reflect the actual video content. This indicates poor encoding of the video. Inspired by~\cite{bai2018empirical_tcn}, we adapt their idea of a temporal convolution network architecture (TCN) to replace the recurrent neural network encoder. The TCN approach was used for sequence to sequence generation on tasks where input consists of long sequences of action segmentation or copy memory. We adapt the TCN model to generate a single output - the embedding for the entire video - in order to provide full information to the decoder before it starts to generate sentences. The decoder structure is the same as before.

The idea behind TCN (Figure \ref{fig:arhitecturi} c) is to capture how features change over time by using one dimensional temporal filters. By employing a hierarchy of convolutions with increasing dilation rate, the amount of information combined increases exponentially, over different time scales, until it reduces the temporal dimension to one, to capture global content. 

For each one of the $N_t$ time steps, we have $N_{fe}$ features, resulting in a tensor of $1 \times N_t \times N_{fe}$ dimension. The network is composed of several blocks of convolution, without padding, in order to reduce the temporal dimension of input from $N_t$ to $1$. Each block has 2 dilated convolutional layers~\cite{yu2015multi}. Each applies several 1x3 filters with Relu nonlinearity and batch normalization~\cite{ioffe2015batch}. The dilation rate is increased with the depth, in order to compute over different scales. Between 2 successive levels, there are residual connections~\cite{he2016deep}. Batch normalization is applied to ease the optimization process. Based on this architecture, we trained several models, varying the network depth, the size of the filters and the dilation rate.

\section{Multiple networks consensus} 
\label{ensamble}

While our models reach a level of accuracy that stands well against published literature, there is a relatively high degree of variation in their output sentences due to the different ways we encode the video content. Some models tend to have complex, descriptive results with a richer vocabulary, while others generate simple, concise sentences. There is also variation in terms of content vs. form. Some sentences are more fluent and complex (e.g. "a man in a suit and a woman talking about the history of the world"), while others are simpler, but better rooted in actual video content (e.g. "a man is talking about a historical topic").
 
We noticed that the group of sentences very often contains correct sentences. In order to validate this observation, we selected for each video, from all sentences generated, the one with the best CIDEr metric~\cite{vedantam2015cider} with respect to ground truth sentences. It turned out that the best selected sentence per video gave on average, over the whole test video set results that are well above state of the art
(Table \ref{table:human}). Then we made another observation: models generally produce sentences that gravitate around the correct meaning.
Thus, noisy sentence variations could be eliminated if the ensemble of networks could work jointly, as a whole. 

Here we propose an efficient \textbf{consensus algorithm} for selecting the best sentence in the group, composed of two stages - a first consensus stage using simple agreements between sentences and a second stage that involves training an Oracle network.  

\paragraph{First consensus stage.}
If the group of sentences generated by the models pool contains a strong cluster united around the correct meaning, then we could find the best sentence as the one which agrees most with the others. Thus, for each sentence we compute its CIDEr score against the others and select the one with the highest score. This idea is relatively simple, yet statistically powerful. If the better sentences form a strong cluster and the weaker ones depart from human annotations in random noisy ways, then we could find the good ones by measuring the level of agreement between each sentence and the remaining ones. Accidental agreements are very rare, while agreements based on good content are more likely. This is the basis of our approach. By selecting the sentence with highest agreement score we maximize our chances of selecting a good sentence. In experiments, selecting the top scored sentence significantly improves the results (Table \ref{table_feat} and Figure \ref{fig:features_vs_architectures}). 

\paragraph{Second consensus stage - Oracle network.}
Often the best quality sentence is within the top $C$ ($C=3$) according to the consensus score at phase 1. 
We added a second level of selection by training an Oracle network to help picking the better sentence at the top. We train the Oracle Net to pick the better of two sentences, given a reference video.  The video encoding consist of the average over frame features. The two sentences are encoded by LSTMs with shared parameters. All features are then concatenated and passed through 3 fully connected layers to obtain the final output. At inference time, we compare each sentence in the top C, selected through phase one consensus as described previously to all of the others. Each sentence is scored based on the number of pairwise victories. We rank all the sentences according to this score and pick the top one.

We train this model on pairs of sentences generated by our models on videos from training set. 
Pairs could include two sentences of the reference video, or one for the reference and the other randomly chosen from other videos. The ground truth label is picked according to the CIDEr score w.r.t the corresponding human annotations on the training set. 

The consensus  algorithm is now complete and proceeds as  follows: 1) For each sentence in the group, compute its CIDEr score against the others; 2) Keep the top-C scored sentences. 3) Re-rank the top C using the Oracle Net and output the top scored sentence.

\section{Experimental analysis}
\label{ablation}

We trained our models on the challenging MSR-VTT 2016 video captioning dataset and benchmark~\cite{MSRVTT_xu2016msr}. This is the main dataset used for experimental testing in recent literature. It contains 10k videos with diverse visual content. Each clip is 10 to 30 seconds long and is annotated with 20 sentences from different people. For comparison with state-of-the-art we use four of the evaluation metrics most often used in the current literature for natural language tasks, shown here in inverse chronological order of their publication date: CIDEr \cite{vedantam2015cider}, METEOR \cite{METEOR}, ROUGE \cite{Lin:2004} and BLEU \cite{Papineni:2002:BMA:1073083.1073135}. 

\paragraph{Models in the pool:}
Our final consensus network works over a pool of 16 models based on the 4 main architectures, differing in the visual, audio and video features used and the number of layers of depth. We observed that the more models we added to the pool the better the performance. Thus, we have 1 basic seq2seq model, 2 seq2seq models
with attention, 2 Two-Wings models, 1 Two-Stage network, 4 TCN models, 4 seq2seq  models with different groups of extra features added to the encoding, 1 seq2seq model with inception features extracted from small patches on a grid, and the last 1 model uses 2 convolutional layers, one over time dimension and the other over feature dimension as an encoder. Our models were trained using Adam optimizer algorithm with a learning rate set to 0.001 and decayed with a factor of 10. The video embedding dimension is set to 512 units and the decoder hidden dimension varies according to additional features provided.

\subsection{Features vs. Network architectures}
Here we present in detail the features we bring in, by concatenating them to the video encoding. We added features incrementally, in 3 phases (\ref{table_feat}).

In the \textbf{initial phase} we use object category features from Inception v3 (type A features) to encode the video frames, with no additional features concatenated to the encoding. In the \textbf{second phase}, we add extra C3D-Resnet~\cite{hara3dcnns} features trained for Kinetics~\cite{kay2017kinetics} action recognition in video and audio features computed from the means and standard deviations of MFCC feature signals extracted from temporal audio segments (type B features). In the \textbf{third phase} of our experiments we bring in stronger higher-level audio features. These are 128-dimension VGG-style deep audio features \cite{AudioVGG} trained on Youtube-70M. Audio features are constructed concatenating averages over five overlapping segments of video. We also added visual features that we trained for multi-words prediction on a large dataset that combines a subset of Youtube8M~\cite{DBLP:journals/corr/Abu-El-HaijaKLN16} and the MSR-VTT videos using as word labels the intersection between their vocabularies. The actual features used are those from the layer preceding the final output. The features added in the third phase are of the type C features. 
 
\newcommand{\rpm}{\raisebox{.2ex}{$\scriptstyle\pm$}}
\newcommand{\STAB}[1]{\begin{tabular}{@{}c@{}}#1\end{tabular}}
\newcommand{\specialcell}[2][c]{%
  \begin{tabular}[#1]{@{}c@{}}#2\end{tabular}}

\newcolumntype{N}{>{\centering\arraybackslash}m{.99in}}
\newcolumntype{M}{>{\centering\arraybackslash}m{.7in}}

\begin{table}[t]
\begin{center}
\begin{tabular}{|l||c|c||c|c||c|c|}
\hline
& \multicolumn{2}{N||}{Group A} & \multicolumn{2}{N||}{Group A+B} & \multicolumn{2}{N|}{Group A+B+C} \\

\hline
 \multicolumn{1}{|c||}{\bf{Model}}  &  \multicolumn{1}{c|}{\specialcell{\bf{Cider}}} &  \multicolumn{1}{c||}{\specialcell{\bf{Meteor} }} &  \multicolumn{1}{c|}{\specialcell{\bf{Cider}}} &  \multicolumn{1}{c||}{\specialcell{\bf{Meteor} }} &  \multicolumn{1}{c|}{\specialcell{\bf{Cider} \\ }} &  \multicolumn{1}{c|}{\specialcell{\bf{Meteor}}}\\ \hline

 Seq2Seq & 36.0 & 25.5 & 44.0 & 27.4 & 46.1 & 28.3\\[0.5ex]

 Two-Wings & 32.2 & 25.2 & 42.2 & 27.3 & 46.2 & 28.8 \\[0.5ex]
  Two-Stage & 34.9 & 25.2 & 43.3 & 27.4 & 45.7 & 28.4\\[0.5ex]

  TCN & 36.80 & 25.5  & 43.9 & 27.4 & 46.1 & 28.4 \\[0.5ex]

  Attention &41.0 & 26.6 & 44.2 & 27.5  & 46.4 & 28.5   \\[0.5ex]
 \hline
   MEAN & 36.0 \rpm  2.5 & 25.6 \rpm 0.5 & 43.9 \rpm 0.6 & 27.4 \rpm 0.2 & 46.0 \rpm 0.9 & 28.4 \rpm 0.3\\[0.5ex]
 \hline\hline

\multicolumn{5}{|c||}{Best individual model}  & 46.2 & 28.8  \\[0.5ex]
\hline
 \multicolumn{5}{|c||}{Consensus}  & 52.1 & 29.6 \\[0.5ex]
 \multicolumn{5}{|c||}{Consensus + OracleNet}  & \bf{53.8} & \bf{29.7}   \\[0.5ex]

\hline
\end{tabular}
\end{center}
    \caption{Performance of our models using different image, video and audio features added during three experimental phases: 
Group A - Inception features; Group B - C3D + MFCC audio features; Group C - VGG audio + Y8M word labels features. In each phase we report the average results of each type of models and the average of all the models. Note how additional features pretrained on different tasks significantly improve performance. Also note the very large performance gain obtained through consensus.}
\label{table_feat}
\end{table}

Our experiments (Table \ref{table_feat} and Figure \ref{fig:features_vs_architectures}) show that the additional, complementary high level information brought in by features pre-trained on different tasks 
have an impact on performance comparable to the consensus procedure. This fact strongly suggests that the intermediate level of semantics captured by these features is important for better bridging the gap between vision and language.  
At the same time, the results suggest that, 
while a great effort has been put into creating video captioning datasets, they are still limited for learning such a challenging task. As we discuss in Section \ref{sec:evaluation_metrics}, another potential limitation comes from the current evaluation metrics used in the literature that seem better at evaluating good sentence form and fluency than at capturing the more profound meaning of sentences.

\subsection{Comparison with the top models}

\begin{figure}[t!]
\centering
\includegraphics[width=.99\textwidth]{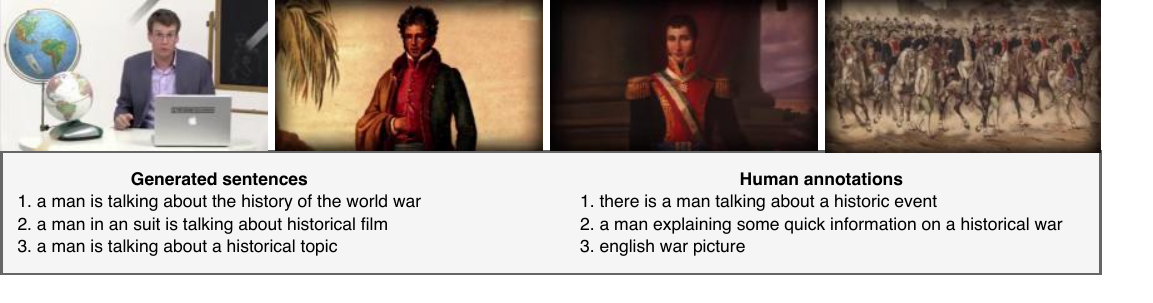}
\caption{Qualitative results showing 3 generated sentences from our models along with a few relevant human annotations. The generated sentences are fluent and related in content to the human annotations. Also note how diverse the human annotations are, especially in form, while being highly meaningful. For more qualitative results please see our project page: \url{https://sites.google.com/view/mining-for-meaning}}
\label{qualitative_results}
\end{figure}

In Table \ref{table: soa} we compare our method against the top submissions from the MSR-VTT 2016 competition, but also against top models published after the competition on that dataset. While our individual models are very competitive in comparison to the top published methods (\ref{table_feat}), the consensus between all models significantly improves the performance achieving state of the art results on several metrics.
For qualitative results of our system please see Figure \ref{qualitative_results} and the supplementary material.

Given the multiple architectures in our system, our encodings are diverse and also are more constrained toward the language space by forcing the additional tasks of language reconstruction or multi-label word prediction. In contrast, other methods use a single encoding scheme such as seq2seq or simply mean pooling the features. The authors of \cite{pasunaruRamBnMoh} use a model similar to our Two-Wings model but they need additional annotated data to learn the language encoding.
In \cite{JinShizheJia}, they use an implicit ensemble by building a decoder defined by a linear mix of parameters conditioned on multiple latent topics. Because of the non-linearity of the language we argue that it is non-trivial to combine words predictions at the model level, and a simple selection between generated sentence improves results.

\begin{table}[t]
\begin{minipage}[b]{0.56\linewidth}

\begin{table}[H]
\begin{center}
\centering
\begin{tabular}{|l|c|c|c|c|c|}
\hline
\bf {}  & \bf {Cider} & \bf {Meteor} & \bf {Rouge} & \bf {Bleu 4} \\[0.05ex]
\hline\hline

\bf{VideoLAB} \cite{ramanishka2016multimodal_videolab}& 44.1 & 27.7 & 60.6 & 39.1 \\[0.05ex]

\bf{v2t navig} \cite{jin2016describing} & 44.8 & 28.2 & 60.9 & 40.8 \\[0.05ex]

\bf{Aalto} \cite{shetty2016frame} & 45.7 & 26.9 & 59.8 & 39.8\\[0.05ex]

\bf{ruc-uva} \cite{dong2016early_ruc}& 45.9 & 26.9  & 58.7 & 38.7\\[0.05ex]

\bf{MT-Ent} \cite{pasunaruRamBnMoh} & 47.1 & 28.8 &  60.2  & 40.8\\[0.05ex]

\bf{HRL} \cite{Wang2017VideoCV} & 48.0 &  28.7 & 61.7  & 41.3 \\[0.05ex]

\bf{dense} \cite{shen2017weakly_video_captioning} & 48.9 & 28.3  & 61.1 &  41.4 \\[0.05ex]

\bf{CIDEnt-RL} \cite{PasRamBanMoh} & 51.7 & 28.4  & {61.4} & {40.5}\\[0.05ex]

\bf{TGM} \cite{JinShizheJia} & 52.9 & \bf{29.7}  & {-} & \bf{45.4}  \\[0.05ex]




\bf{Ours} & \bf{53.8} & \bf{29.7} & \bf{63.0} & 44.2   \\[0.05ex]

\hline 
\end{tabular}
\end{center}
    \caption{Comparison with the top models on MSR-VTT 2016 test dataset. We obtain state of the art results on three evaluation metrics. 
    }
\label{table: soa}
\end{table}

\end{minipage}\hfill
\begin{minipage}[b]{0.32\linewidth}

\begin{figure}[H]
\includegraphics[scale=0.055]{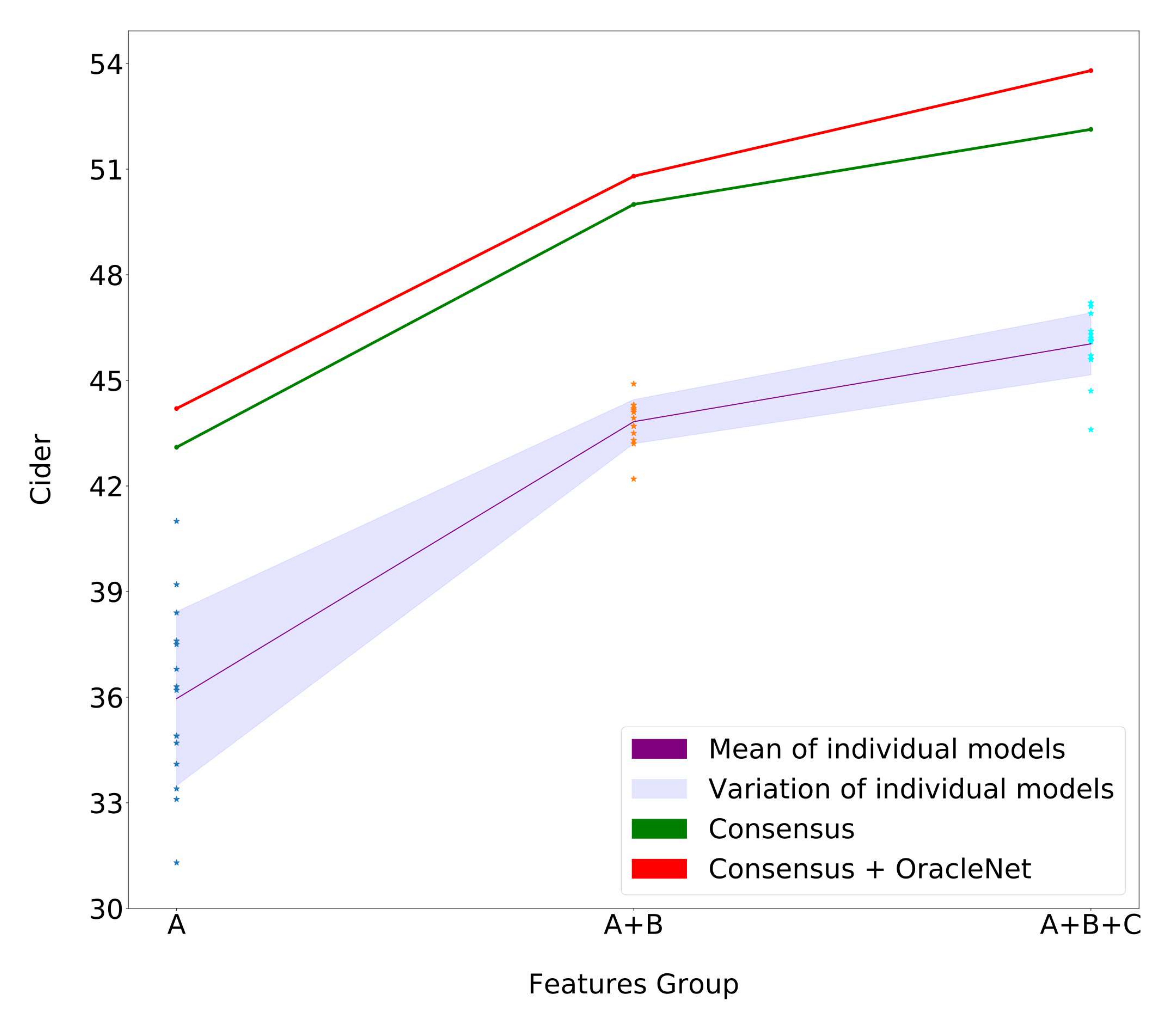}
\vspace{-1ex}
\caption{Mean and std of CIDEr score for all our individual models over the 3 features phases along with consensus networks performance.}
\vspace{-5ex}
\label{fig:features_vs_architectures}
\end{figure}

\label{fig:image}
\end{minipage}
\end{table}

\subsection{A short discussion on language understanding and evaluation}
\label{sec:evaluation_metrics}

On close inspection, we found that very small changes in the sentence structure or at the level of words, without changing the overall meaning, may strongly change the metric scores. Consequently, we measure how humans perform against each other on MSR-VTT (Table~\ref{table:human})(by computing the metrics between one human annotation versus the rest) and observed the same instability in the metric scores. The human agreement in terms of the different metrics is quite low and often below the performance of our system (Table \ref{table:human}). 
Since it is evident that our system does not \emph{speak} better than a human,  it must be the metrics that are not quite appropriate. Current methods for sentence generation might be in fact very close to a certain saturation of these metrics.  

The truth is that designing good evaluation metrics is, in this case, almost as hard as the research task itself.
How could we automatically evaluate 
the hidden meaning of a sentence if we have not learned yet how to encode this meaning? This seems like a chicken and egg problem. The "meaning" is usually hidden, so we are almost forced to evaluate form, which is explicit. However, there are infinitely many correct sentence forms for a given video sequence.  
One could expect that a common, higher level representation for understanding the story of what happens in the scene is needed, which sits above vision and language form. Our work suggests that such a representation could be learned indirectly, in a  distributed way through intermediate high level but simpler tasks, such as the detection of actions, interactions or more abstract entities (word-labels). They could sit above the physical objects but still below full language expression. 

\begin{table}[ht]
\begin{center}
\begin{tabular}{|l|c|c|c|c|c|}
\hline
\bf {}  & \bf {Cider} & \bf {Meteor}  & \bf {Rouge} & \bf {Bleu} \\[0.5ex]
\hline\hline

\textbf{Human avg} & 50.2 \rpm 6.5 & 29.8 \rpm 3.5  & 73.1 \rpm 5.4  & 34.5 \rpm 8.2 \\[0.5ex]
\textbf{Human worst} & 4.0  & 15.2  & 31.9  & 7.4  \\[0.5ex]
\textbf{Human best} & 108.1 & 48.1 & 84.0 & 76.1  \\[0.5ex]
\hline
\textbf{Ours worst} & 18.8 & 22.0 & 51.1  & 24.0  \\[0.5ex]
\textbf{Ours best} & 75.3 & 34.6 & 69.7 & 55.3  \\[0.5ex]
\hline
\textbf{Ours}  & 53.8 & 29.7 & 63.0 & 44.2  \\[0.5ex]
\hline 
\end{tabular}
\end{center}
\caption{Human performance for MSR-VTT test dataset. Human worst/best are computed by selecting for every video the worst/best sentence with respect to the rest. Ours worst/best are computed by selecting for every video the worst/best scores with respect to ground truth. }
\label{table:human}
\end{table}

In this work we argue that it could be possible to select more meaningful sentences if we look for the consensus of many networks which have learned to output fluent linguistic descriptions that are rooted in powerful visual and audio features pretrained on simpler-to-evaluate, yet high level semantic tasks (such as action or words prediction). Populations of such networks might reach a shared implicit \emph{meaning} through multiple networks consensus.

\section{Conclusions}

In this work, we presented different architectural approaches to encode content of a video for the purpose of learning to generate captions. We also studied the impact of external features pre-trained on other intermediate tasks and concluded that such features have a strong impact on performance. Describing the dynamic world as it changes through space and time is a very exciting but still extremely challenging problem, which is not well understood. In order to cope with various limitations and reduce the noise of each individual model, we propose a novel approach for
finding high quality sentences using multiple encoder-decoder networks.
We argue that while the evaluation metrics might not be perfect for individual sentence evaluation, we could still reach correct meaningful sentences through the statistically powerful multiple networks consensus algorithm. We clearly demonstrate the value of our approach by achieving state of the art results on the challenging MSR-VTT benchmark.

\paragraph{Acknowledgements:} The authors would like to thank Elena Burceanu and Ioana Croitoru for helpful comments and discussions. This work was partially supported by UEFISCDI, under project PN-III-P4-ID-ERC-2016-0007 and  PN-III-P2-2.1-PED-2016-1842.

\bibliography{egbib}

\section{Appendix}

In this Section we present some additional qualitative results. Firstly, we give some results produced by our pool of models and show the top sentences when they are ranked by the score with respect to the ground truth versus when they are ranked by the consensus score. Ideally, the consensus score should produce a ranking as close as possible to the ranking produced by the comparison to the ground truth. We then present results from the language reconstruction part of the Two-Wings Network and show that it is effective in producing coherent sentences from unordered sets of words. In the final part we display the intermediate, predicted word labels by the Two-Stage Network to better understand their relation to the final caption produced.

\paragraph{Consensus vs. Ground Truth Ranking:}
In Tables \ref{table_res2}, \ref{table_res3}, \ref{table_res4} and \ref{table_res5} we show some qualitative examples of sentences generated by our 16 models from multiple videos. On the left column of the tables we present 4 frames sampled from each video. The right side is split in 3 cells: in the upper cell we present top 5 generated sentences sorted by consensus score (with actual value shown at the start of each line), in the middle cell we list the top 5 sentences ordered by CIDER score with respect to ground truth (actual value shown at the start of each line) and in the bottom cell we randomly sample 5 examples of from human annotations. 

Notice that our models produce meaningful and coherent sentences.
In general we observed that the rank of a sentence in the order given by the consensus score is strongly correlated with the true rank in the order given by the ground truth score. This fact can also be seen in the examples presented below. Thus, the top scoring consensus sentences are also top scores with respect to ground truth. Therefore the consensus is a powerful automatic ranking scheme that could be reliably used to select good quality, meaningful sentences. The top-C captions are likely to contain top sentences with respect to ground truth, from which the Oracle Network can make the final selection. Also note that the human annotations, while generally agreeing on content, have varied degrees of fluency and quality.

\begin{table}[h!]
\caption{Example results generated by the pool of our 16 models. We present sentences ordered by consensus (first row) and by CIDEr score computed w.r.t. ground truth (second row) and human annotation (final row).}

\resizebox{1.0 \textwidth}{!}{   
\begin{tabular}{|l|l|l|}
\hline

\makecell{ 
\includegraphics[width=0.30\textwidth]{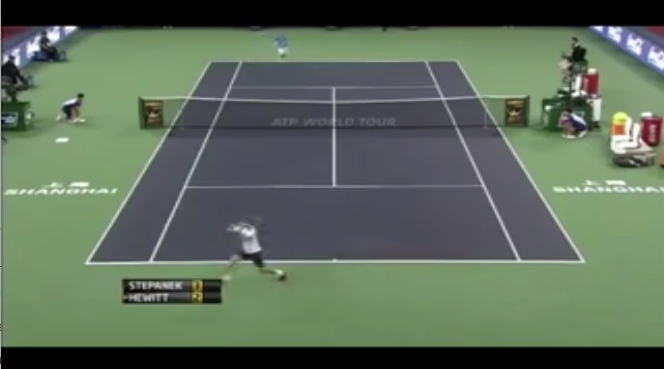} \\
\includegraphics[width=0.30\textwidth]{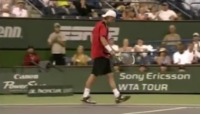} \\ 
\includegraphics[width=0.30\textwidth]{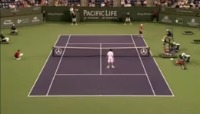} \\ 
\includegraphics[width=0.30\textwidth]{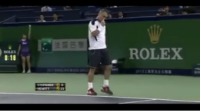} 
} 

&
\makecell[l]{
\textbf{Top generated sentences ordered by consensus scores} \\
2.798. two men are playing tennis on a court\\
2.451. two men are playing tennis on a tennis court\\
2.065. a tennis player in blue and blue shirt is playing tennis\\
2.041. two men are playing a tennis game\\
1.938. a tennis player is hitting a ball in a match\\
\hline
\textbf{Top generated sentences ordered by CIDEr score w.r.t. ground truth} \\
1.670. two men are playing tennis on a court \\
1.577. a tennis match is being played between two players\\
1.443. two men are playing a tennis game\\
1.356. two men are playing tennis on a tennis court\\
1.296. a tennis match between two men in blue and blue shirt\\
\hline
\textbf{Human annotations} \\
a tennis match is being played between two men\\
two men participate and play in a tennis match\\
two people playing in a tennis match\\
a tennis match between two men with an advertisement for rolex\\
a tennis piont ends with one player s signature shot\\
}\\[0.5ex]
\hline

\makecell{
\includegraphics[width=0.30\textwidth]{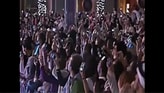} \\
\includegraphics[width=0.30\textwidth]{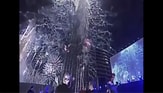} \\ 
\includegraphics[width=0.30\textwidth]{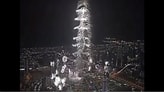} \\ 
\includegraphics[width=0.30\textwidth]{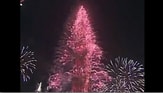} 
} 
&
\makecell[l]{
\textbf{Top generated sentences ordered by consensus score} \\
3.287. a group of people are watching a fireworks display\\
3.179. a large crowd of people are watching a fireworks display\\
2.731. a crowd of people are watching a fireworks show\\
2.689. a large fireworks display is being shown\\
2.440. a large fireworks display\\
\hline
\textbf{Top generated sentences ordered by CIDEr score w.r.t. ground truth} \\
1.536. a group of people are watching a fireworks display\\
1.527. a crowd of people are watching a fireworks show\\
1.330. a group of fireworks are going to the crowd\\
1.325. a large crowd of people are watching a fireworks display\\
1.209. a fireworks display is going off\\
\hline
\textbf{Human annotations} \\
a clip showcasing fireworks going off in the sky\\
a crowd is cheering at fireworks\\
a crowd of people are recording a fireworks show with their cellphones\\
a crows takes photos of fireworks shooting from a building\\
a group of people are watching fireworks
}\\[0.5ex]
\hline
\end{tabular}
} 

\label{table_res2}
\end{table}

\begin{table}[h!]
\caption{Additional qualitative results of sentences ranked by consensus versus ground truth ranking. }
\resizebox{1.0 \textwidth}{!}{   
\begin{tabular}{|l|l|l|}
\hline

\makecell{ 
\includegraphics[width=0.30\textwidth]{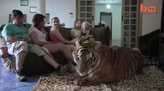} \\
\includegraphics[width=0.30\textwidth]{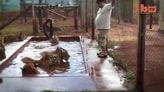} \\ 
\includegraphics[width=0.30\textwidth]{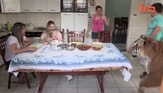} \\ 
\includegraphics[width=0.30\textwidth]{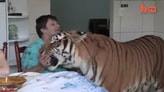} 
} 
&
\makecell[l]{
\textbf{Top generated sentences ordered by consensus score} \\
2.321.  a man and a woman are sitting in a table\\
1.118.  a man is eating food from a table\\
1.070.  a man is sitting in a table with a big dog\\
1.069.  a group of people are sitting in a line with a tiger\\
0.925.  a man is sitting in a chair with a tiger\\
\hline
\textbf{Top generated sentences ordered by CIDEr score w.r.t. ground truth} \\
0.650.  a group of people are sitting in a line with a tiger\\
0.599.  a man is sitting in a chair with a tiger\\
0.588. a man and a woman are eating a tiger in a bowl\\
0.547.  a man is talking about a tiger\\
0.180.  a man and a woman are sitting in a table\\
\hline
\textbf{Human annotations} \\
a story about a family that has seven tigers\\
a family rearing tigers and feeding them in the home\\
a family and their children are sitting at a table playing with a tiger\\
five people sitting on a couch and a tiger laying by their feet'\\
a family in brazil has 7 tigers that live in the house as petst\\
}\\[0.5ex]
\hline

\makecell{ 
\includegraphics[width=0.30\textwidth]{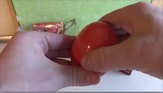} \\
\includegraphics[width=0.30\textwidth]{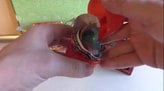} \\ 
\includegraphics[width=0.30\textwidth]{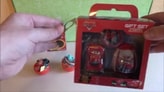} \\ 
\includegraphics[width=0.30\textwidth]{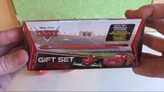} 
} 
&
\makecell[l]{
\textbf{Top generated sentences ordered by consensus score} \\
5.699.  a person is opening a toy\\
4.238.  a person is opening a toy with a toy\\
4.039.  a person is opening a package\\
4.015.  a person is opening a box\\
3.666.  a person is opening a red package \\
\hline
\textbf{Top generated sentences ordered by CIDEr score w.r.t. ground truth} \\
2.131.  a person is opening a toy\\
1.586.  a person is opening a box\\
1.445.  a person is opening a toy with a toy\\
1.441.  a person is opening a package\\
1.367.  a person is opening a red package\\
\hline
\textbf{Human annotations} \\
a clip of someone taking toys out of a gift set \\
a man is opening a toy egg\\
a man is playing with some toys\\
a man is showing off the cars gift set\\
an unboxing of some toys\\
}\\[0.5ex]
\hline


\hline
\end{tabular}

} 

\label{table_res3}
\end{table}

\begin{table}[h!]
\caption{Additional qualitative results of sentences ranked by consensus versus ground truth ranking. }
\resizebox{1.0 \textwidth}{!}{   
\begin{tabular}{|l|l|l|}
\hline

\makecell{ 
\includegraphics[width=0.30\textwidth]{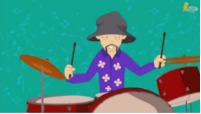} \\
\includegraphics[width=0.30\textwidth]{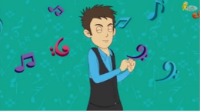} \\ 
\includegraphics[width=0.30\textwidth]{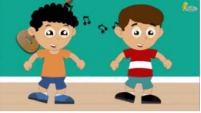} \\ 
\includegraphics[width=0.30\textwidth]{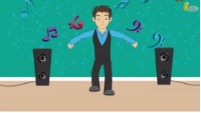} 
} 
&
\makecell[l]{
\textbf{Top generated sentences ordered by consensus score} \\
3.271.  a cartoon character is singing\\
2.997.  a cartoon character is singing a song\\
2.727.  a cartoon of a man singing a song\\
2.705.  a cartoon of a man singing and dancing\\
2.682.  a cartoon of a girl singing and dancing \\
\hline
\textbf{Top generated sentences ordered by CIDEr score w.r.t. ground truth} \\
0.646.  a cartoon of a man singing and dancing\\
0.627.  a cartoon of a girl singing and dancing\\
0.479.  a cartoon is singing\\
0.450.  a cartoon character is dancing\\
0.423.  a cartoon of a man singing a song\\
\hline
\textbf{Human annotations} \\
a  kid  s animated song \\
a animation band is singing song\\
a cartoon about the hokey pokey song and dance\\
a cartoon depicts the hokey pokey\\
a cartoon for children\\
}\\[0.5ex]
\hline

\makecell{ 
\includegraphics[width=0.30\textwidth]{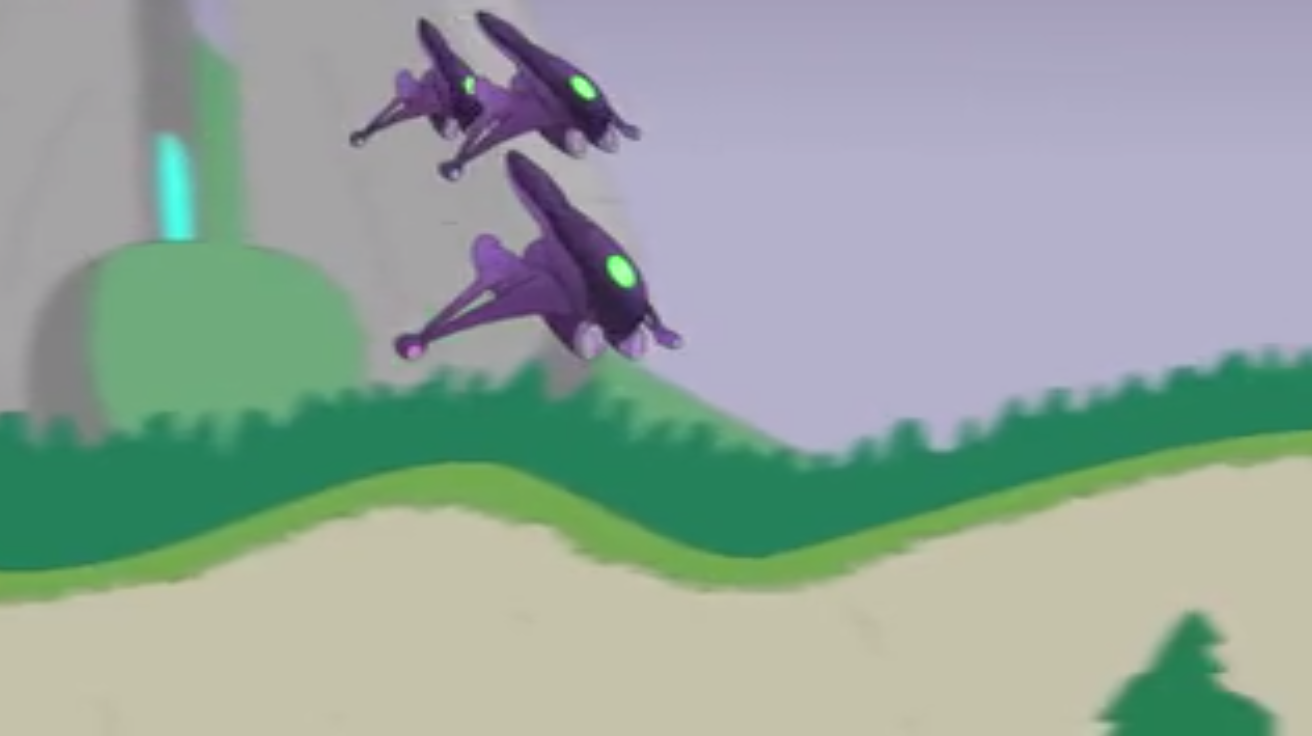} \\
\includegraphics[width=0.30\textwidth]{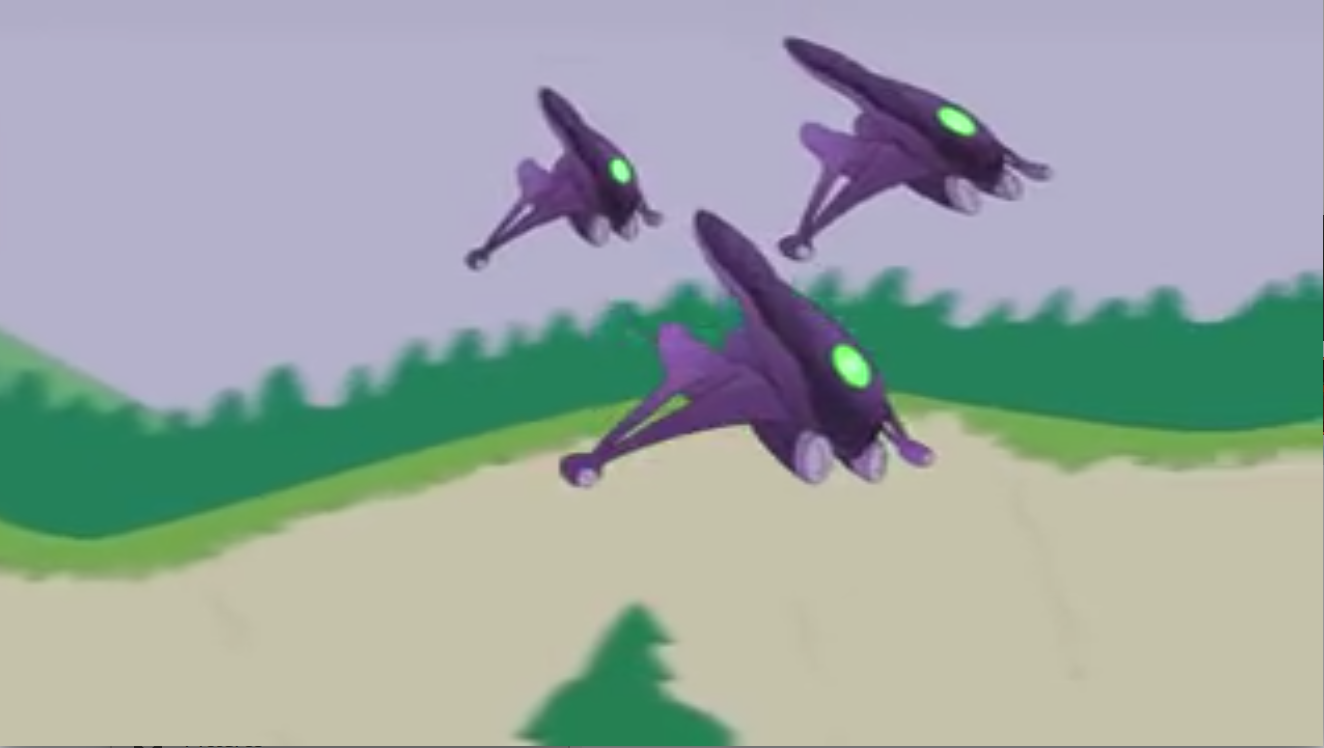} \\ 
\includegraphics[width=0.30\textwidth]{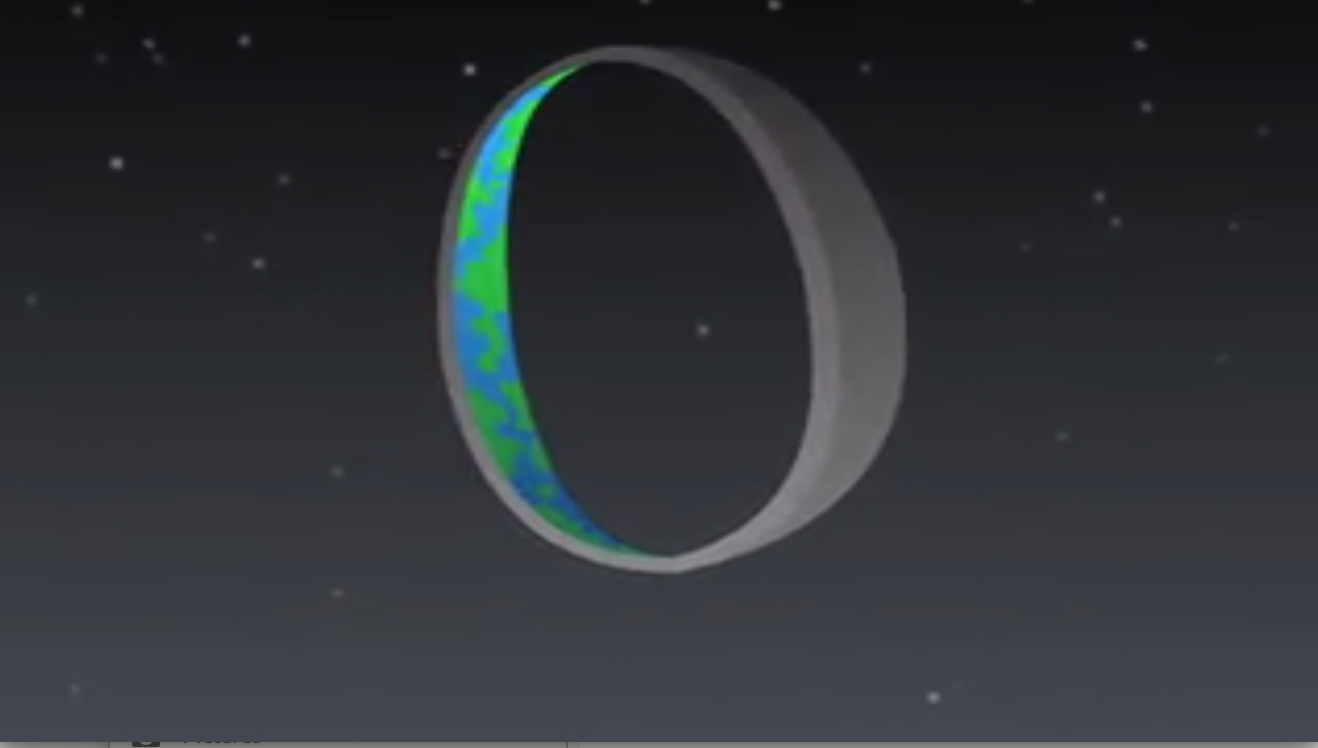} \\ 
\includegraphics[width=0.30\textwidth]{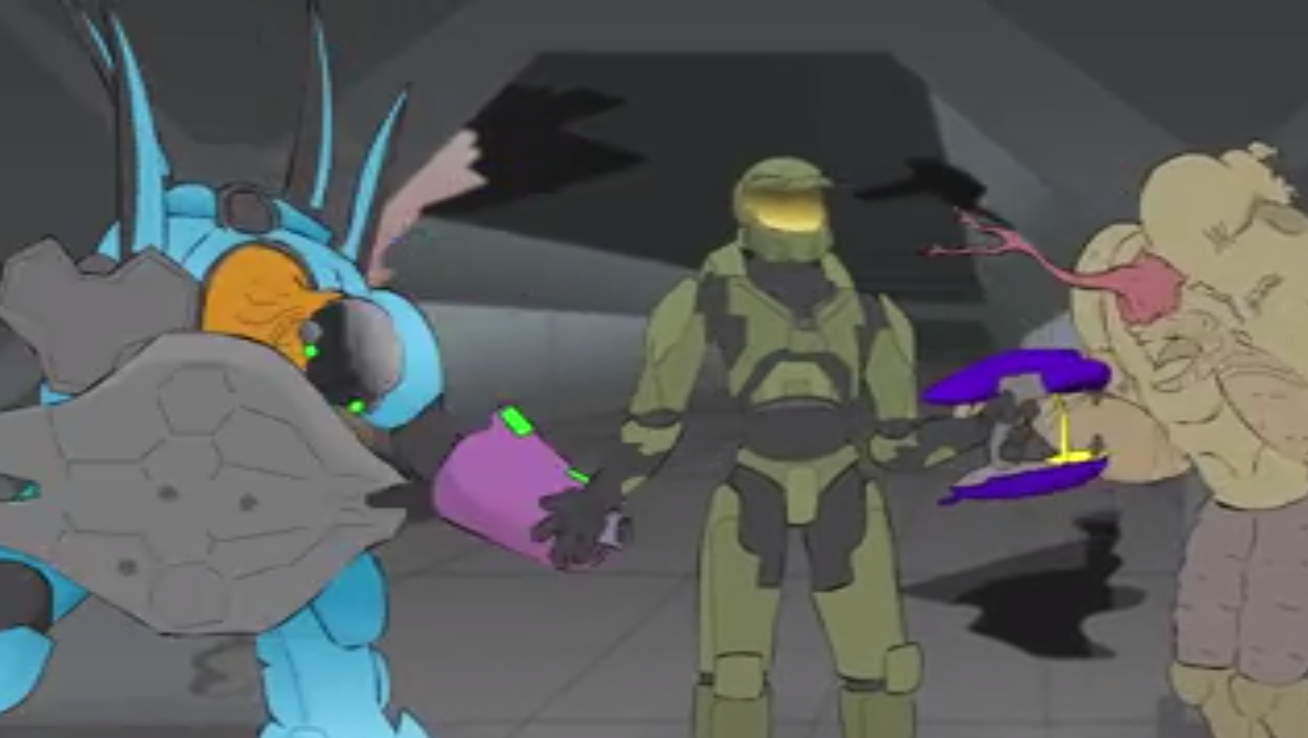} 
} 
&
\makecell[l]{
\textbf{Top generated sentences ordered by consensus score} \\
3.380.  a video game character is flying \\
3.170.  a video game character is flying around \\
2.493.  a cartoon character is flying \\
2.388.  a cartoon character is flying a ball \\
2.268.  a person is playing a video game \\
\hline
\textbf{Top generated sentences ordered by CIDEr score w.r.t. ground truth} \\
0.314.  a cartoon of a man is flying around \\
0.269.  a cartoon character is flying \\
0.243.  a man is flying through a video game \\
0.233.  a cartoon character is flying a ball \\
0.202.  a cartoon character flying a monster \\
\hline
\textbf{Human annotations} \\
video game characters are flying through space\\
there were three characters flying in the air\\
cartoon characters sing about space\\
a video of halo the video game\\
halo cartoon animation music video\\
}\\[0.5ex]
\hline


\hline
\end{tabular}

} 

\label{table_res4}
\end{table}

\begin{table}[h!]
\caption{Additional qualitative results of sentences ranked by consensus versus ground truth ranking. }
\resizebox{1.0 \textwidth}{!}{   
\begin{tabular}{|l|l|l|}
\hline

\makecell{ 
\includegraphics[width=0.30\textwidth]{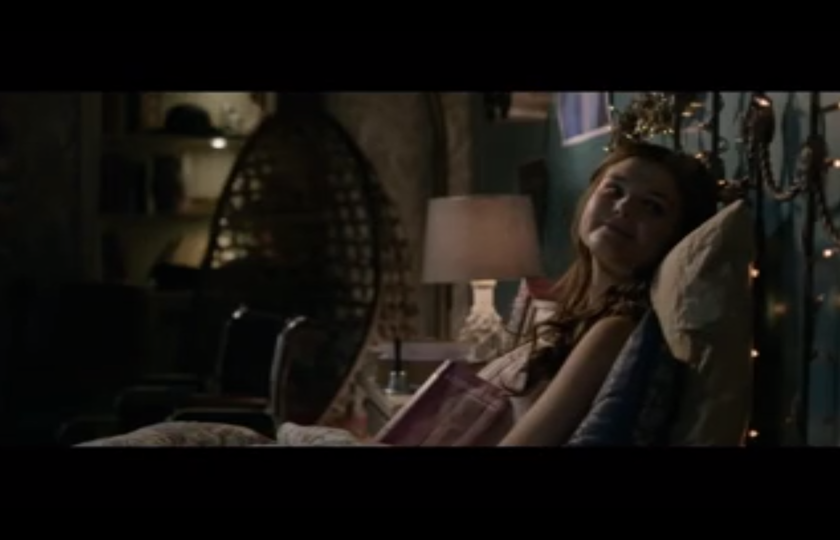} \\
\includegraphics[width=0.30\textwidth]{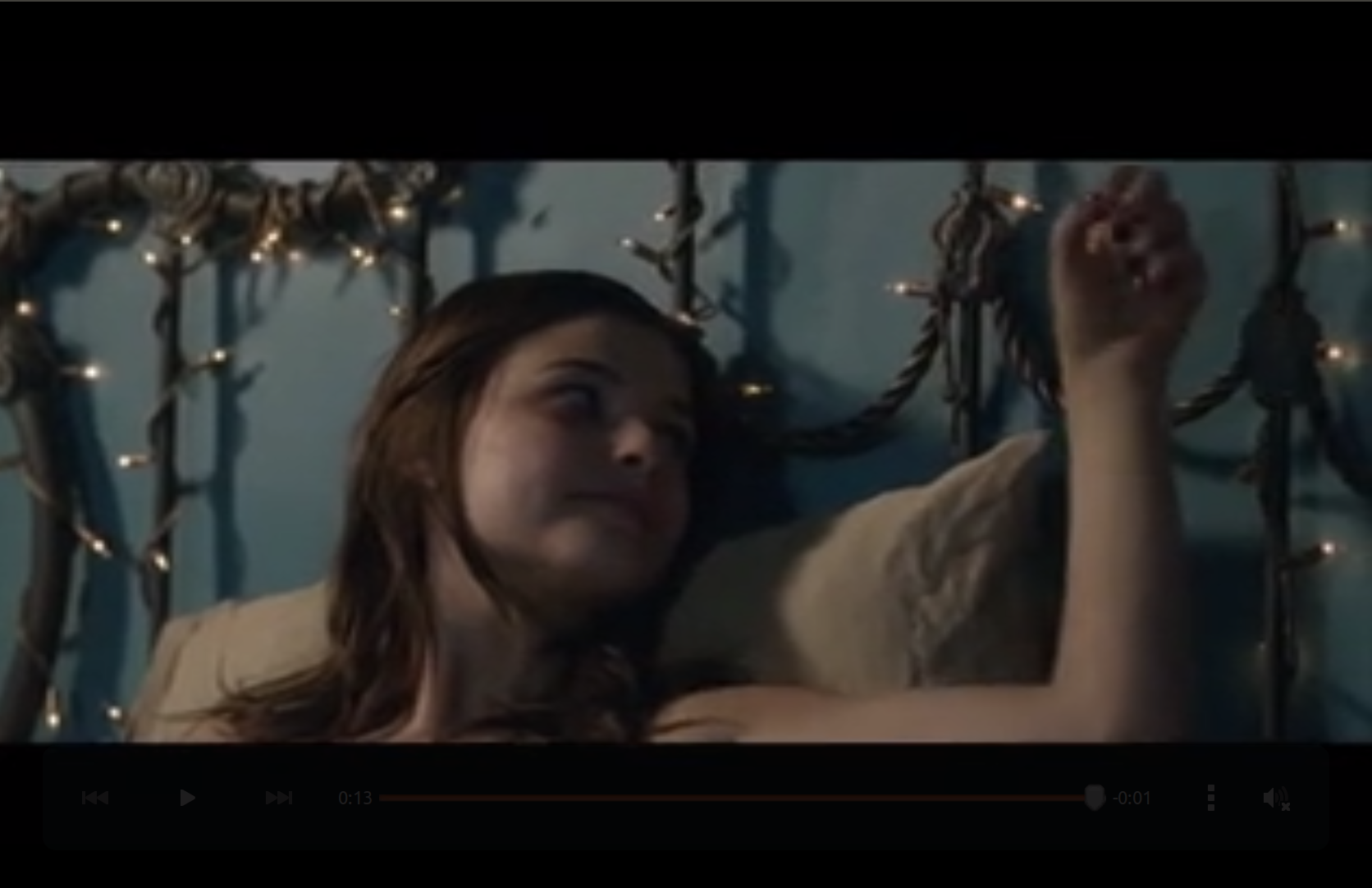} \\ 
\includegraphics[width=0.30\textwidth]{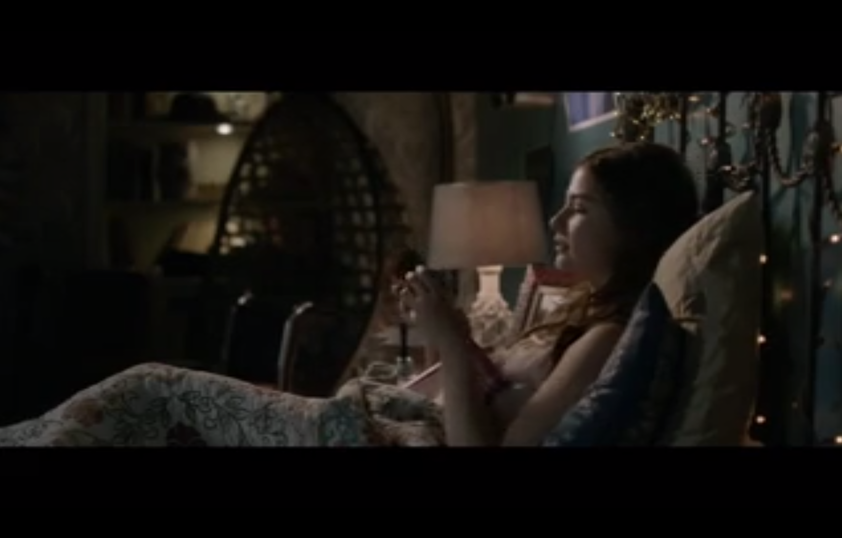} \\
\includegraphics[width=0.30\textwidth]{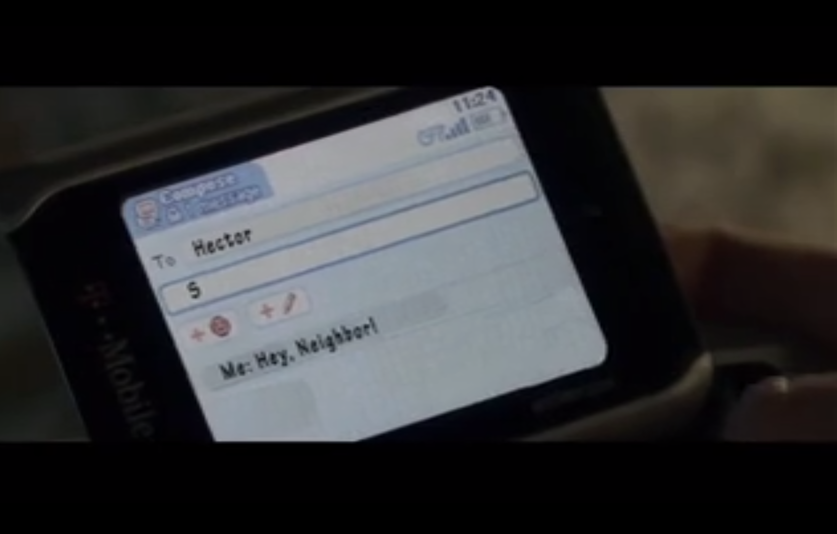} \\
} 
&
\makecell[l]{
\textbf{Top generated sentences ordered by consensus score} \\
4.660.  a girl is knocking on the wall and texting \\
4.103.  a girl is knocking on the wall \\
3.601.  a girl is knocking on a wall and texting \\
3.359.  a girl is knocking on her phone \\
3.119. a girl is knocking on her bed \\
\hline
\textbf{Top generated sentences ordered by CIDEr score w.r.t. ground truth} \\
1.518.  a girl is knocking on the wall \\
1.315.  a girl is knocking on the wall and texting \\
1.310.  a girl laying in bed and knocking on the wall \\
1.289.  a girl is laying in bed and knocking on the wall \\
1.109.  a girl is knocking on a wall and texting \\
\hline
\textbf{Human annotations} \\
a girl in bed \\
a girl is knocking on the wall \\
a girl knocking on a wall \\
a girl knocks on a wall and texts a friend \\
a girl lays in bed and uses her phone \\
}\\[0.5ex]
\hline
\makecell{ 
\includegraphics[width=0.30\textwidth]{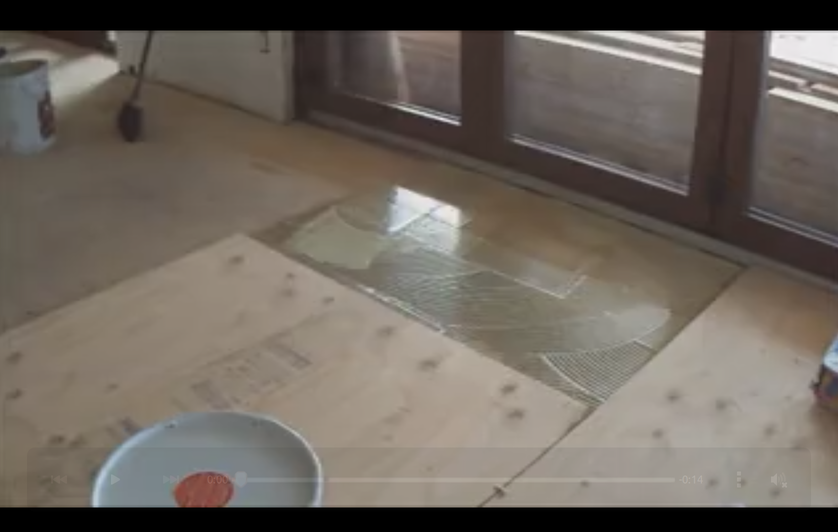} \\
\includegraphics[width=0.30\textwidth]{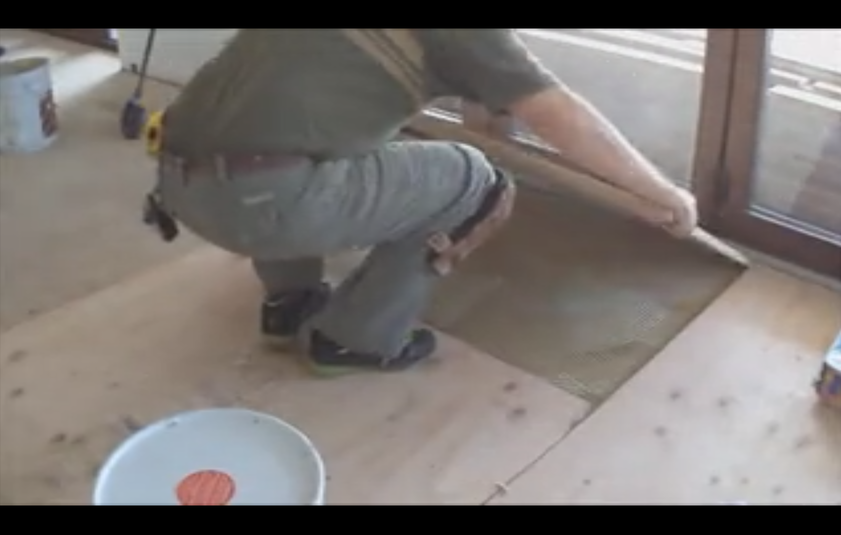} \\ 
\includegraphics[width=0.30\textwidth]{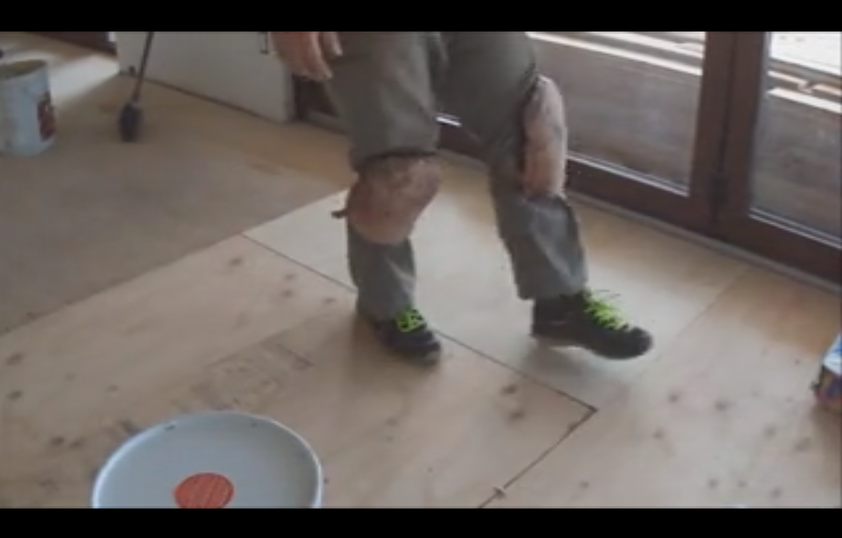} \\ 
\includegraphics[width=0.30\textwidth]{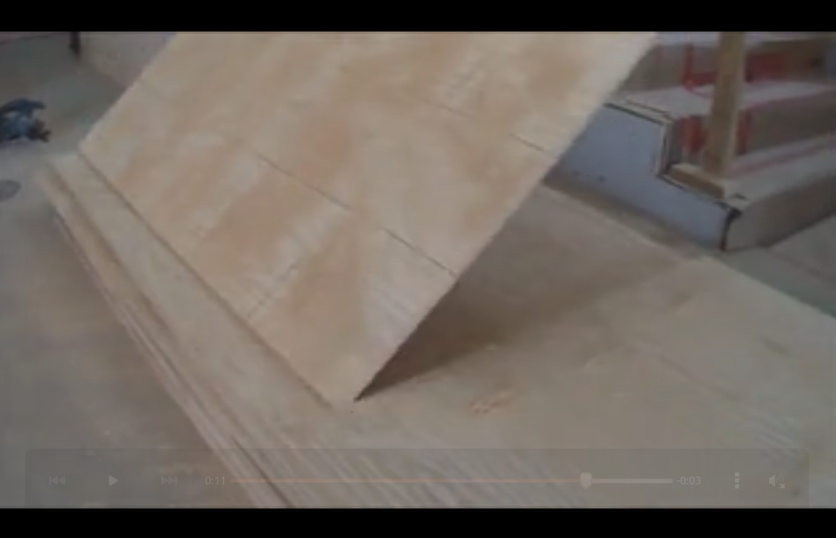} 
} 
&
\makecell[l]{
\textbf{Top generated sentences ordered by consensus score} \\
1.991.  a man is doing construction \\
1.907.  a man is doing construction work \\
1.881.  a man is doing construction improvement \\
1.855.  a man is working on a floor \\
1.842. a man is installing a wood floor \\
\hline
\textbf{Top generated sentences ordered by CIDEr score w.r.t. ground truth} \\
1.132. a man is installing a wood floor \\
1.052.  a man is installing flooring \\
0.867.  a man is working on a wood floor \\
0.816.  a man is fixing a wood floor \\
0.621.  a man is doing flooring \\
\hline
\textbf{Human annotations} \\
a  man is installing new flooring \\
a carpenter places down some wood floring \\
a man is decking a floor \\
a man is fixing the floor \\
a man is flooring \\
}\\[0.5ex]
\hline
\hline
\end{tabular}
} 

\label{table_res5}
\end{table}

\clearpage
\paragraph{Language reconstruction results:}
In Table \ref{tab:reconstruction_results} we show results of the reconstruction part of the Two-Wings model. In this submodel we receive as input a sentence from the annotations, apply a random permutation on the order of the words, remove half of them and try to reconstruct it. The first column contains the sentence to be reconstructed, the second  column the remaining shuffled words used as input and the third one the generated sentence. 

We can see that the generated captions are grammatically and semantically correct. Although the reconstruction does not match the target sentence, given just half of the individual words and the random permutation of these words it would be impossible, even for a human, to reconstruct back the original sentence.
But this is not the end goal of this branch - its main purpose is to learn to generate rich, diverse and coherent sentences.
Consequently, in our experiments the Two-Wings Network produced on average more diverse sentences than other network models.

\begin{table}[h!]
\caption{ Results of the language reconstruction part of the Two-Wings model. The first column contains the ground truth sentence, the second the broken input sentence and the last the generated sentence. We use this model to improve our language generating decoder.}
\label{tab:exTable1}

\makebox[1 \textwidth][c]{       
\resizebox{1.0 \textwidth}{!}{   
\begin{tabular}{|c|c|c|c|}
\hline
\rowcolor{Gray}
\textbf{Ground truth Sentence} & \textbf{Input words} & \textbf{Generated Sentence}   \\[0.5ex]
\hline
\makecell{ a man talks about a young \\ girl performing great in track} & 
\makecell{ a about great performing \\ young man a talks girl in} & 
\makecell{ a man talking about a young \\ girl in front of a crowd} \\[0.5ex]
    \hline
\makecell{ someone giving demo in \\ computer about something  } & 
\makecell{  in about computer someone \\ demo giving } & 
\makecell{ someone is showing video \\ graphics on computer } \\[0.5ex]
\hline
\makecell{ a boy in a long sleeve shirt \\ is playing the guitar} & 
\makecell{a boy the in long is sleeve a \\  shirt playing} & 
\makecell{a young boy is playing in the \\  water with a dog} \\[0.5ex]
\hline
\makecell{ a man is performing on stage \\ with some other actors} & 
\makecell{ stage with some on \\  performing is man a} & 
\makecell{ a man is singing \\  with his band on stage} \\[0.5ex]
\hline
\makecell{two guy: one in black tshirt \\is driving a car and other \\ in white shirt is sitting next to him} & 
\makecell{is in and in sitting other car \\guy white tshirt  two is a shirt\\ driving black one } & 
\makecell{ a man in a blue shirt is sitting\\ and speaking with two other \\ men while standing on a couch
} \\[0.5ex]

\hline
\makecell{ a balding man in a suit \\ gives a speech by flags} & 
\makecell{ speech balding a suit \\ in a a man gives} & 
\makecell{  a man in a suit with \\ a tie talking in a room } \\[0.5ex]
\hline
\makecell{ a baseball player in a red \\ uniform while music plays} & 
\makecell{ player in music red  \\ uniform a a baseball while} & 
\makecell{ a man in a blue shirt and \\ black shorts playing basketball} \\[0.5ex]

\hline
\end{tabular}

} 
} 

\label{tab:reconstruction_results}
\end{table}

\begin{figure}[H]
\centering
\includegraphics[width=.99\textwidth]{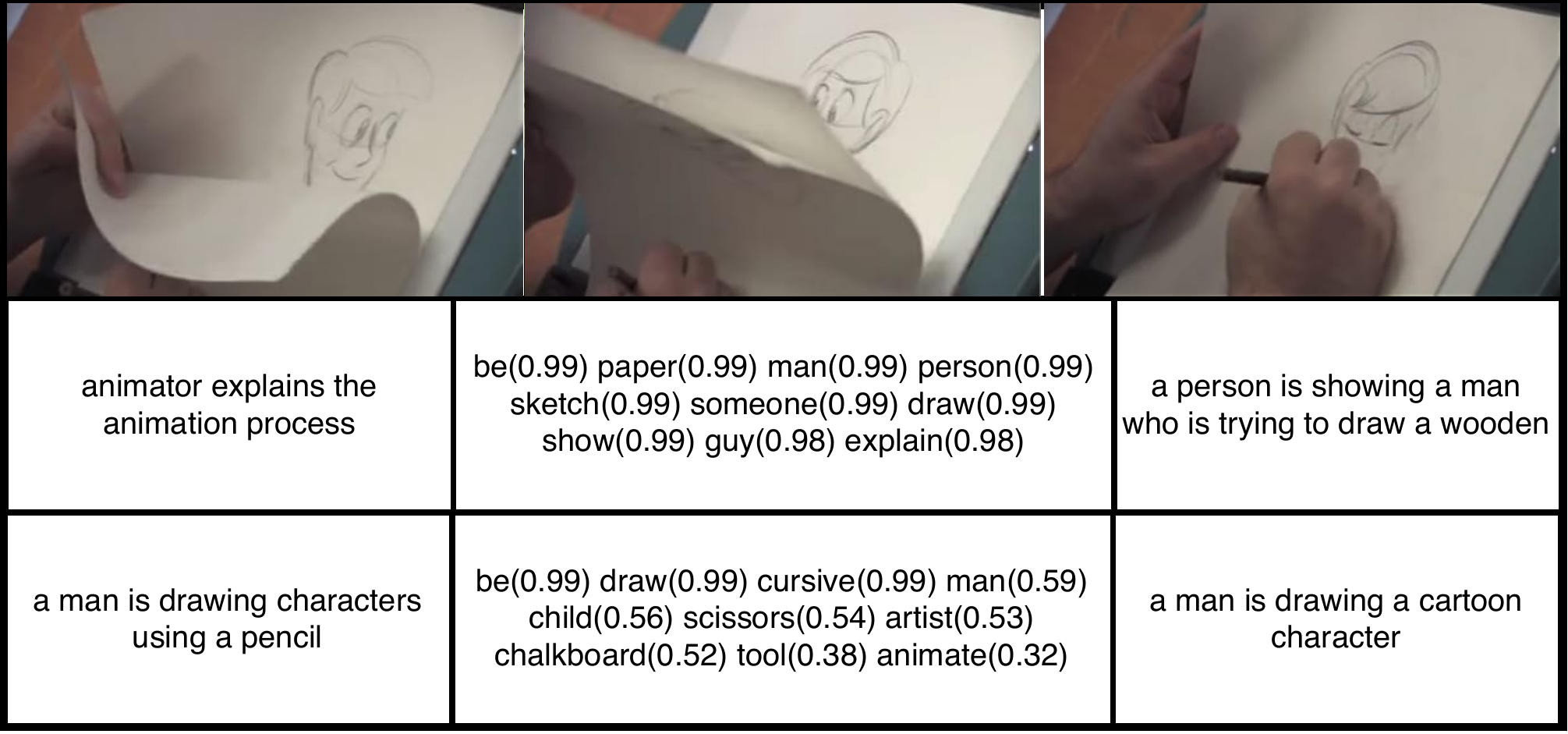}
\includegraphics[width=.99\textwidth]{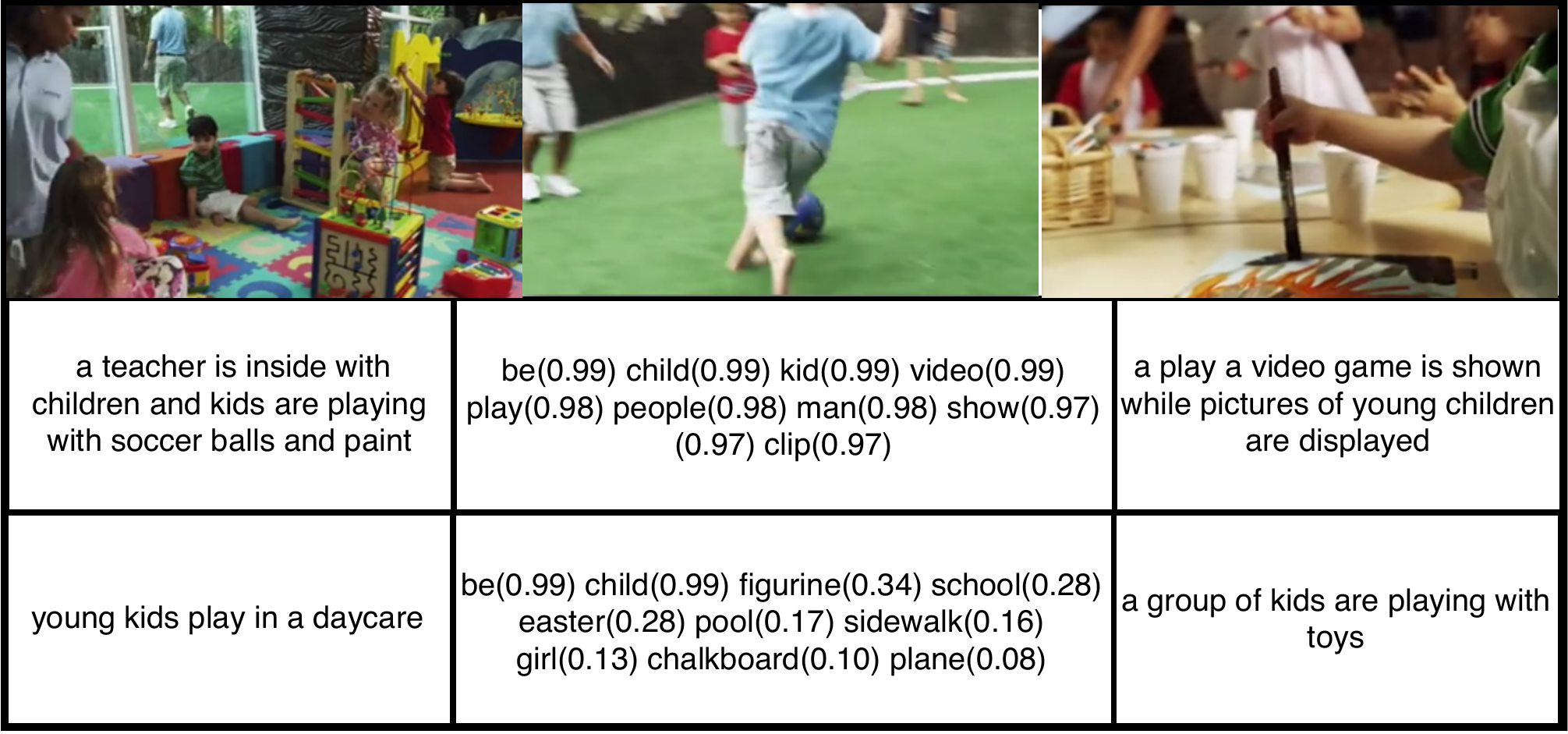}
\includegraphics[width=.99\textwidth]{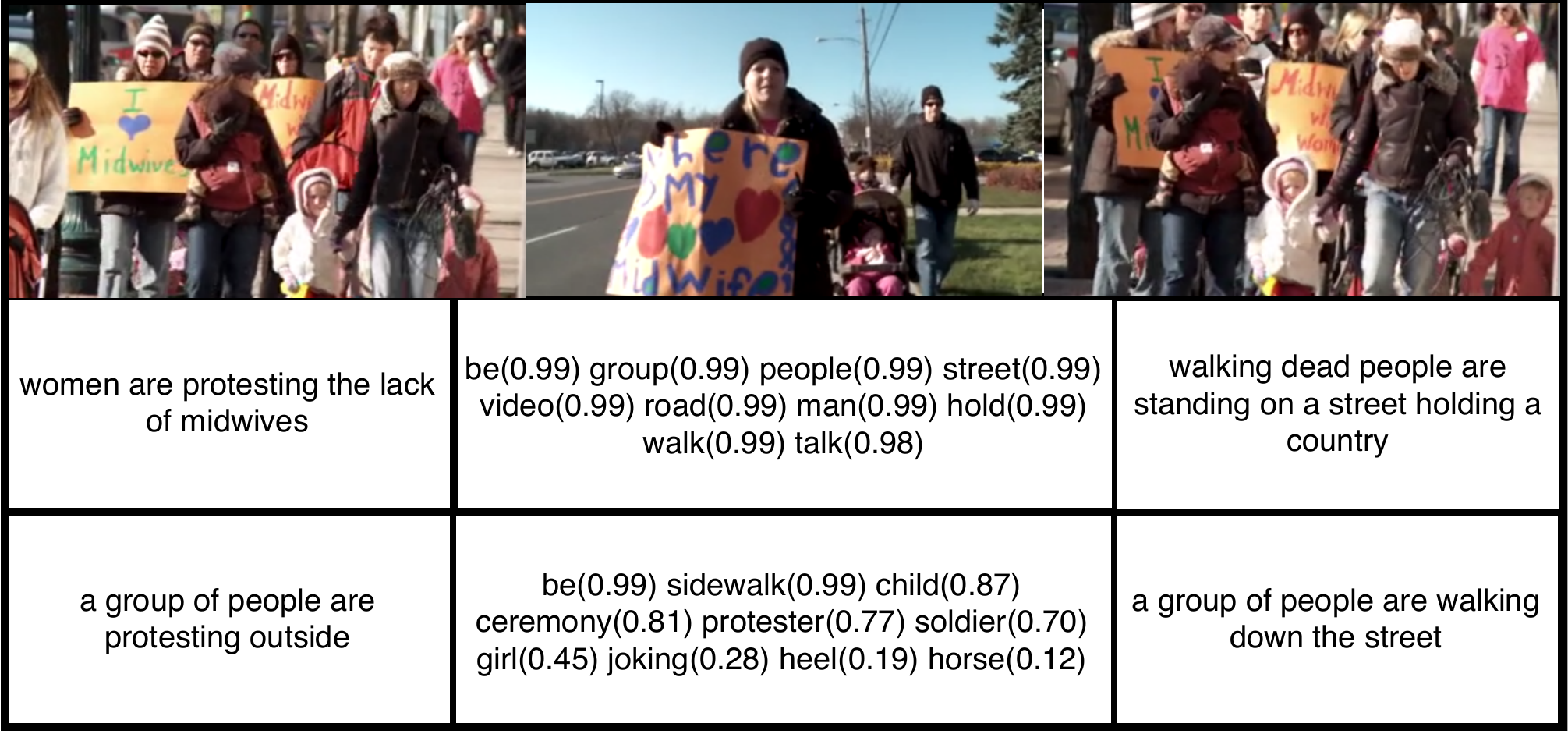}

\caption{Example of qualitative results of the Two-Stage Network. The first column contains a ground truth sentence, the second column the top 10 predicted labels and the third column contains the final generated sentence by the Two-Stage model. In top row of every box, we present the results of the model when we train the two parts separately. In the second row we show results of the two parts fine-tuned jointly, end-to-end on the MSR-VTT dataset.}

\label{rezultate_two_stage}
\end{figure}

\paragraph{Two-Stage Network results:}
In Figure \ref{rezultate_two_stage} we present some examples produced by the Two-Stage Model. The model consists of two parts: a part (a first stage) that predicts multiple labels for the video followed by a part (the next stage) that generates a sentence based only on these labels. The model is initialized by training both parts independently and then fine-tune them jointly.

For each cell, the first row contains 3 frames sampled from each video, the second row contains results from the initial independent learning and the third row contains final results after the fine-tuning. For each row, the first column shows a ground truth sentence, the second column shows the top K predicted labels with their corresponding probabilities and the last column shows the generated sentence.


Labels generated by our multi-label model have a high degree of accuracy. To improve the quality of the captions we fine-tuned the whole model end-to-end, obtaining significantly better results. While the multi-label prediction was better before fine-tuning, after the fine-tuning, which did not put a cost on the labels, the multi-label prediction decreased in accuracy, while the final quality at the caption level improved. This fact is also observed in the qualitative examples in Figure \ref{rezultate_two_stage} as the fine-tuned model, trained end-to-end (third row) produces captions of better quality than the model with the two stages trained independently (second row). However, the fine tuned model is worse at predicting intermediate word labels - due to the end-to-end training with loss on the final caption but no intermediate loss on the intermediate multi-label prediction.

\end{document}